%% file: main.tex
\documentclass[lettersize,journal]{IEEEtran}
\usepackage{amsmath,amsfonts}
\usepackage{algorithmic}
\usepackage{algorithm}
\usepackage{array}
\usepackage[caption=false,font=normalsize,labelfont=sf,textfont=sf]{subfig}
\usepackage{textcomp}
\usepackage{stfloats}
\usepackage{url}
\usepackage{verbatim}
\usepackage{graphicx}
\usepackage{cite}
\usepackage{array}
\usepackage[utf8]{inputenc} 
\usepackage[T1]{fontenc}    
\usepackage{tcolorbox}
\definecolor{cadmiumgreen}{rgb}{0.0, 0.42, 0.24}
\definecolor{oldmauve}{rgb}{0.4, 0.19, 0.28}
\definecolor{royalazure}{rgb}{0.0, 0.22, 0.66}
\definecolor{harvardcrimson}{rgb}{0.79, 0.0, 0.09}
\definecolor{lightmauve}{rgb}{0.86, 0.82, 1.0}
\definecolor{darkbrown}{rgb}{0.4, 0.26, 0.13}%
\usepackage[colorlinks = true,
            linkcolor = harvardcrimson,
            urlcolor  = harvardcrimson,
            citecolor = royalazure,
            anchorcolor = royalazure]{hyperref}
\usepackage{url}           
\usepackage{booktabs}       
\usepackage{amsfonts}       
\usepackage{nicefrac}      
\usepackage{microtype}     
\usepackage{bbding}
\usepackage{microtype}
\usepackage{graphicx}
\usepackage{booktabs} 
\usepackage{times,epsfig}
\usepackage{bm}
\usepackage{amsfonts,amssymb,amsmath,amsthm}
\usepackage{threeparttable}
\usepackage{array}
 \usepackage{multirow}
\usepackage{threeparttable}
\usepackage{wasysym}
\usepackage{extarrows}
\usepackage{longtable}
\usepackage{supertabular}
\usepackage{amsmath}
\usepackage{amssymb}
\usepackage{algorithm}
\usepackage{algorithmic}
\usepackage{makecell}
\usepackage{pifont}
\usepackage{enumerate}
\usepackage{tikz}
\usetikzlibrary{decorations.pathmorphing}
\usetikzlibrary{trees}
\definecolor{mygreen}{RGB}{18, 119, 51}
\definecolor{myred}{RGB}{178, 25, 45}
\usepackage{makecell}

\usepackage{xcolor}
\long\def\comment#1{}
\def\ie{$i.e.$}
\def\eg{$e.g.$}

\def\wrt{$w.r.t.$}
\def\etal{$et~al.$}

\usepackage{amsthm}
\usepackage{arydshln}

\theoremstyle{definition}

\usepackage{color, colortbl}
\definecolor{Gray}{gray}{0.9}
\definecolor{LightCyan}{rgb}{0.88,1,1}

\usepackage{enumitem}

\def\x{\mathbf{x}}

\usepackage{tikz}
\usepackage{tikz}

\definecolor{color1}{rgb}{0.1,0.1,0.1}
\definecolor{color2}{rgb}{0.2,0.2,0.2}  

\makeatletter
\newcommand\bigDiamond{\mathop{\mathpalette\bigDi@mond\relax}}
\newcommand\bigDi@mond[2]{%
  \vcenter{\hbox{\m@th
    \scalebox{\ifx#1\displaystyle 2\else1.2\fi}{$#1\Diamond$}%
  }}%
}
\newcommand\bigLozenge{\mathop{\mathpalette\bigL@zenge\relax}}
\newcommand\bigL@zenge[2]{%
  \vcenter{\hbox{\m@th
    \scalebox{\ifx#1\displaystyle 2\else1.2\fi}{$#1\blacklozenge$}%
  }}%
}
\makeatother

\usepackage{booktabs}

\long\def\comment#1{}
\usepackage{xcolor}

\usepackage{multirow}
\usepackage{booktabs} 
\usepackage{graphicx}

\usepackage{stackengine}
\usepackage{scalerel}

\usepackage{fontawesome}

\begin{document}

\title{Defenses in Adversarial Machine Learning: A Survey}

\author{
Baoyuan Wu, Shaokui Wei, Mingli Zhu, Meixi Zheng, Zihao Zhu, Mingda Zhang, \\Hongrui Chen, Danni Yuan, Li Liu, Qingshan Liu
\\
\IEEEcompsocitemizethanks{
\IEEEcompsocthanksitem 
The first eight authors are with the School of Data Science, The Chinese University of Hong Kong, Shenzhen, China, email: wubaoyuan@cuhk.edu.cn, shaokuiwei@link.cuhk.edu.cn, minglizhu@link.cuhk.edu.cn, meixizheng@link.cuhk.edu.cn, zihaozhu@link.cuhk.edu.cn, mingda zhang@link.cuhk.edu.cn, hongruichen@link.cuhk.edu.cn, danniyuan@link.cuhk.edu.cn.
Li Liu is with the Hong Kong University of Science and Technology (Guangzhou), China, email: avrillliu@hkust-gz.edu.cn.
Qingshan Liu is with Nanjing University of Information Science and Technology, email: qsliu@nuist.edu.cn.
\IEEEcompsocthanksitem 
Corresponding author: Baoyuan Wu (wubaoyuan@cuhk.edu.cn).
}
}



\maketitle

\begin{abstract}
Adversarial phenomenon has been widely observed in machine learning (ML) systems, especially in those using deep neural networks, describing that ML systems may produce inconsistent and incomprehensible predictions with humans at some particular cases. This phenomenon poses a serious security threat to the practical application of ML systems, and several advanced attack paradigms have been developed to explore it, mainly including backdoor attacks, weight attacks, and adversarial examples. For each individual attack paradigm, various defense paradigms have been developed to improve the model robustness against the corresponding attack paradigm.  However, due to the independence and diversity of these defense paradigms, it is difficult to examine the overall robustness of a ML system against different kinds of attacks.
This survey aims to build a systematic review of all existing defense paradigms from a unified perspective. Specifically, from the life-cycle perspective, we factorize a complete machine learning system into five stages, including pre-training, training, post-training, deployment, and inference stages, respectively. Then, we present a clear taxonomy to categorize and review representative defense methods at each individual stage.
The unified perspective and presented taxonomies not only facilitate the analysis of the mechanism of each defense paradigm but also help us to understand connections and differences among different defense paradigms, which may inspire future research to develop more advanced, comprehensive defenses.
\end{abstract}


\input{sections/section-introduction.tex}

\input{sections/section-unified-perspective}

\input{sections/section-pre-processing}

\input{sections/section-training}

\input{sections/section-post-processing}

\input{sections/section-inference}

\input{sections/section-discussion}

\input{sections/section-conclusion.tex}

{\small
\bibliographystyle{ieee_fullname}
\bibliography{reference-defense}
}


 





\end{document}

%% file: sections/section-introduction.tex
\section{Introduction}
\IEEEPARstart{A}{dversarial} vulnerability of machine learning models has been considered as one of the main security threats in practice, which can compromise the integrity, availability, and confidentiality of machine learning systems. To better understand this vulnerability, three primary attack paradigms have emerged, targeting the training, deployment, and inference stages of machine learning systems, respectively referred to as backdoor attacks, weight attacks, and adversarial examples. In response to these various forms of attack, numerous defense strategies have been proposed to reinforce the resistance of machine learning models against one of these attacks. All works of both adversarial attacks and adversarial defenses belong to the topic called \textbf{adversarial machine learning} (AML). For the attack aspect of AML, one latest survey \cite{wu2023adversarial} has presented a unified definition and a general formulation, clearly demonstrating the connections and differences among these three attack paradigms. This work aims to provide a comprehensive review of existing defense methods from a unified perspective, thereby completing the full landscape of AML research.

There have been a few surveys to review the defenses against one particular attack paradigm, such as the defenses against the training-stage attack \cite{gao2020backdoor,goldblum2022dataset}, the defenses against the deployment-stage attack \cite{qian2023survey}, as well as the defenses against the inference-stage attack \cite{qian2022survey,wei2022physically}.  However, the surveys of different defense paradigms often present totally different and rather diverse taxonomies to summarize the covered works, due to the independence and diversity of these defense paradigms. This lack of consistency makes it challenging to obtain a comprehensive view of adversarial defenses across all stages of a machine learning system's lifecycle. Consequently, the performance of each individual defense method could not be comprehensively evaluated, the claimed robustness guarantee may be one-sided or even misleading, and the trade-offs and limitations may not be clearly revealed. For example, adversarial training is effective in defending against inference-time adversarial examples but may be ineffective against the data poisoning-based backdoor attack \cite{weng2020trade, gao2023effectiveness}. 

To address the above limitations, here we attempt to provide a unified perspective on the defense aspect of AML on the image classification task, based on the life-cycle of the machine learning system.
Specifically, instead of categorizing existing defenses solely according to their goals at the first level, we first categorize them by the distinct phases of the machine learning life-cycle, \ie, pre-training stage, training stage, post-training, deployment stage, and inference stage. Subsequently, the corresponding defenses in each stage are further categorized according to their underlying goals.  Such a taxonomy provides two advantages. 
\textbf{First,} it provides a clear alignment between diverse defenses and diverse attacks at different stages. 
\textbf{Second,} it enables comparison and analysis between different defenses belonging to the same stage, to highlight their connections and differences. 
Moreover, we hope that this unified life-cycle perspective and the corresponding taxonomy could facilitate future research on developing more comprehensive defenses throughout the life-cycle of machine learning systems. 

\textbf{Our contributions} In this paper, we make the following contributions to the field of AML defense: \textbf{1)} We propose a unified perspective on the defense aspect of AML, which aligns diverse defense paradigms and attack paradigms based on the stages of machine learning system’s life-cycle.
\textbf{2)} We present comprehensive taxonomies to categorize the defense methods at different stages, and provide a systematic survey of the existing defenses against various attacks. 
\textbf{3)} We identify the open challenges and future directions for developing more effective and robust defenses throughout the life-cycle of the machine learning system.

\textbf{Organization} The remaining contents of the manuscript are organized as follows. Section~\ref{sec:unified} provides a unified perspective of defenses in AML, based on the life-cycle of the machine learning systems.  Section~\ref{sec: defense at pre-training} covers the methods that enhance the data and model robustness at the pre-training stage, preparing the essential components for subsequent training. Section~\ref{sec: defense at training} reviews the methods that enhance the model robustness at the training stage, following which we review the post-training/deployment stage methods in  Section~\ref{sec: defense at post-training}. In Section~\ref{sec: defense at inference}, we investigate the methods that enhance the model robustness at the inference stage. Section~\ref{sec: discussion} discusses the challenges and opportunities in current research, and highlights some applications of these defense methods. Finally, We end with some conclusions in Section~\ref{sec:conclusion}.

%% file: sections/section-unified-perspective.tex
\section{A Unified Perspective of Defenses in Adversarial Machine Learning}
\label{sec:unified}

In this section, we first introduce some common notations and definitions used in AML. After that, we present a unified perspective of defense methods in adversarial machine learning based on the stages of AML life-cycle. 

\subsection{Notations}  
We use the following notations throughout this paper. $f_{\boldsymbol{\theta}}$ is a DNN model with model parameter $\boldsymbol{\theta}$. $\mathcal{D}$ is the set of data $(\x,y)$, where $\x$ is the input data and $y$ is the corresponding label. We use $\boldsymbol{\epsilon}$ to denote the adversarial perturbation, such that $\x+\boldsymbol{\epsilon}$ is the perturbed input, \ie, adversarial example. For backdoor attack, $y^{\prime}$ and $\blacktriangle$ are the target label and the pre-designed trigger, respectively. $\mathcal{L}$ is the loss function (e.g., cross-entropy) and $\hat{y}$ is the predicted label for a given input.

\begin{table}[h]
    \centering
    \caption{Table of Notations}
  \begin{tabular}{c|l}
    \toprule
    Notation         & Description/Definition \\
    \midrule
    $\x,y$ & Original (clean) input data and label \\
    $\mathcal{D}$ & Training data distribution \\
     $f_{\boldsymbol{\theta}}(\cdot)$ & DNN model\\
     $\blacktriangle$ & Pre-designed trigger\\
     $\delta$ & Generated perturbation by adv. algorithm \\
     $\hat{y}$ & Prediction of model\\
     $y^{\prime}$ & Target label of adversarial example\\
     $\boldsymbol{\epsilon}$ & Perturbation constraint \\
     $\mathcal{L}$  & Loss function (\eg, cross-entropy)\\
     $T$  & Transformation to input sample\\
     $\tau$  & Transformation to the gradient of model parameters\\
    \bottomrule
  \end{tabular}
  \end{table}

\begin{figure}
    \centering
    \includegraphics[width=1\linewidth]{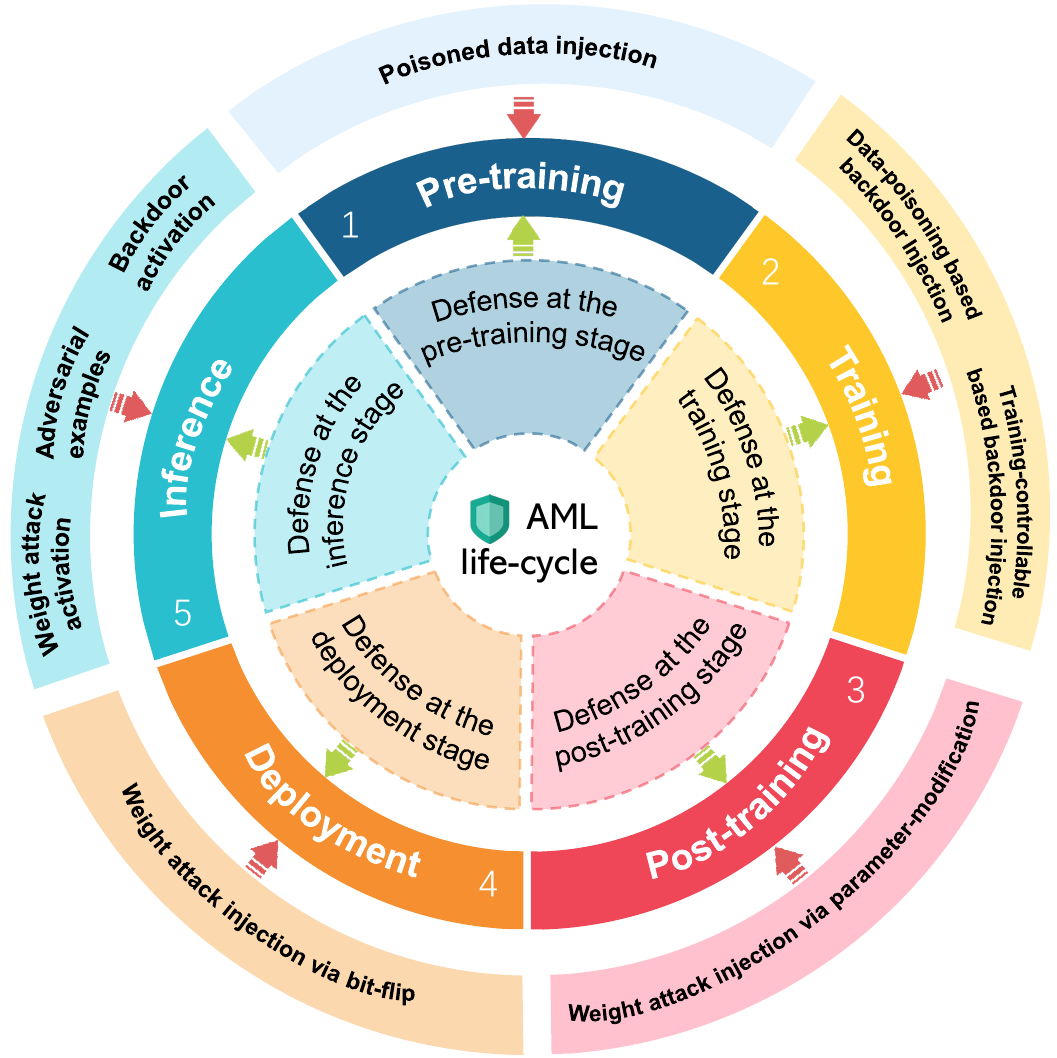}
    \caption{Loop illustration of the life-cycle of AML (the middle loop), as well as the attacks (the outer loop) and the defenses (the inner loop) at each stages.}
    \label{fig:loop structre of AML}
\end{figure}

\input{tables/table-stage-definition}

\subsection{Five stages of the AML life-cycle }
As shown in Fig. \ref{fig:loop structre of AML}, we categorize the defense methods according to the stages in the AML life-cycle, including pre-training, training, post-training, deployment, and inference. Each stage has its own inputs and outputs as summarized in Table~\ref{tab: definitions of terms in AML}, based on which different methods can be developed to defend against specific attacks. We briefly introduce each stage as follows:

\paragraph{Defense at the pre-processing stage} At the pre-processing stage, the defender has control of both the training data and the model architecture. The goal of this stage is to filter out and/or purify the malicious or noisy data, and design robust model architectures, which improves the robustness of model after training.

\paragraph{Defense at the training stage} At the training stage, the defender has full control of the training process, such as the objective function, the optimization algorithm, and the learning paradigm/procedures. The goal of this stage is to enhance the robustness of the model by training it with carefully designed training process.

\paragraph{Defense at the post-training stage} At the post-training stage, only a potentially backdoored model and a small clean dataset (optional) are given, and the main threat is the backdoor attack. To defend against backdoor attacks, one approach is to perform backdoor detection/identification first, after which the model is discarded or purified if it's detected to be poisoned. Another way is to perform backdoor removal/mitigation methods and remove the potential backdoor effect to purify the model.

\paragraph{Defense at the deployment stage} At the deployment stage, the defender has access to a quantized model that is deployed on a device or a platform and the main task is to defend against weight attack. To achieve such a goal, one approach is to enhance the model's resistance to weight attack. Another popular approach is to generate the signature of model and check the signature to detect the weight attack.

\paragraph{Defense at the inference stage} At the inference stage, the defender only has control of the input samples. The goal of this stage is to reject or correct malicious (\eg, adversarial or backdoor) inputs.

%% file: tables/table-stage-definition.tex
\renewcommand\arraystretch{1.2}
\begin{table*}[ht]
\centering
\caption{Categorization of various defenses against various attacks based on AML Lile-cycle.}
\vspace{-0.9em}
\label{tab: definitions of terms in AML}
\scalebox{0.8}{
\begin{tabular}
{m{.08\textwidth}<{\centering} 
 m{.3\textwidth} m{.2\textwidth}
 m{.19\textwidth} m{.3\textwidth}}
\hline 
 Stage & \multicolumn{1}{c}{Task} & \multicolumn{1}{c}{Technology} & \multicolumn{1}{c}{Input} & \multicolumn{1}{c}{Output}
 \\
\hline 
\hline 
 \multirow{3}{*}{Pre-training} & Defend against adv. examples & Robust model architecture & An initial model architecture & A model architecture 
\\
 \cline{2-5}
  & Defend against data-poisoning based backdoor attack & Poisoned data detection & A training dataset & A cleaned dataset and a suspicious dataset
\\
\hline 
 \multirow{5}{*}{Training} 
 & Defend against adv. examples & Adversarial training & A training dataset, a model architecture & A model robust to adv. example 
\\
 \cline{2-5}
& Defend against data-poisoning based backdoor attack & Secure centralized training & A poisoned training dataset, a model architecture & A clean model 
\\
 \cline{2-5}
& Defend against partially training controllable based backdoor attack & Secure decentralized training & A federated learning framework, a controlled server & A clean global model 
\\
\hline 

 \multirow{5}{*}{Post-training} & \multirow{5}{*}{Defend against backdoor attack} & Backdoored model detection & A trained model, a reserved dataset (optional) & The model is backdoored or clean 
\\
 \cline{3-5}
&  & Target label prediction & A trained model, a reserved dataset (optional) & The targeted label(s)
\\
 \cline{3-5}
&  & Backdoor mitigation/removal & A trained model, a reserved dataset (optional) & A clean model 
\\
\hline 
\multirow{3}{*}{Deployment} & \multirow{3}{*}{Defend against weight attack} & Model enhancement & A quantified model,  a training dataset (optional)  & A quantified model with higher fault tolerance 
\\
\cline{3-5} 
&  & Model fingerprint & A quantified model,  a training dataset (optional) & A quantified model with fingerprint 
\\
\hline
\multirow{8}{*}{Inference} & \multirow{3}{*}{Defend against backdoor attack} & Poisoned data detection & A query data, a trained model (optional)  & The query data is poisoned or clean 
\\
\cline{3-5} 
&  & Robust prediction & A query data, a trained model (optional)  & Correct prediction \\
\cline{2-5}
&  \multirow{5}{*}{Defend against adv. examples} & Adversarial data detection & A query data, a trained model (optional) & The query data is adversarial or benign 
\\
\cline{3-5}
&  & Data denoising/purification & A query data, a trained model (optional)  & A denoised/purified data
\\
\cline{3-5}
&  & Dynamic inference & A (sequential) query data, a trained model (optional)  & A feedback with randomness 
\\
\hline
\end{tabular}
}
\end{table*}

%% file: sections/section-pre-processing.tex
\section{Defense at the Pre-training Stage}
\label{sec: defense at pre-training}

\begin{figure*}
    \centering
    \includegraphics[width=1\linewidth]{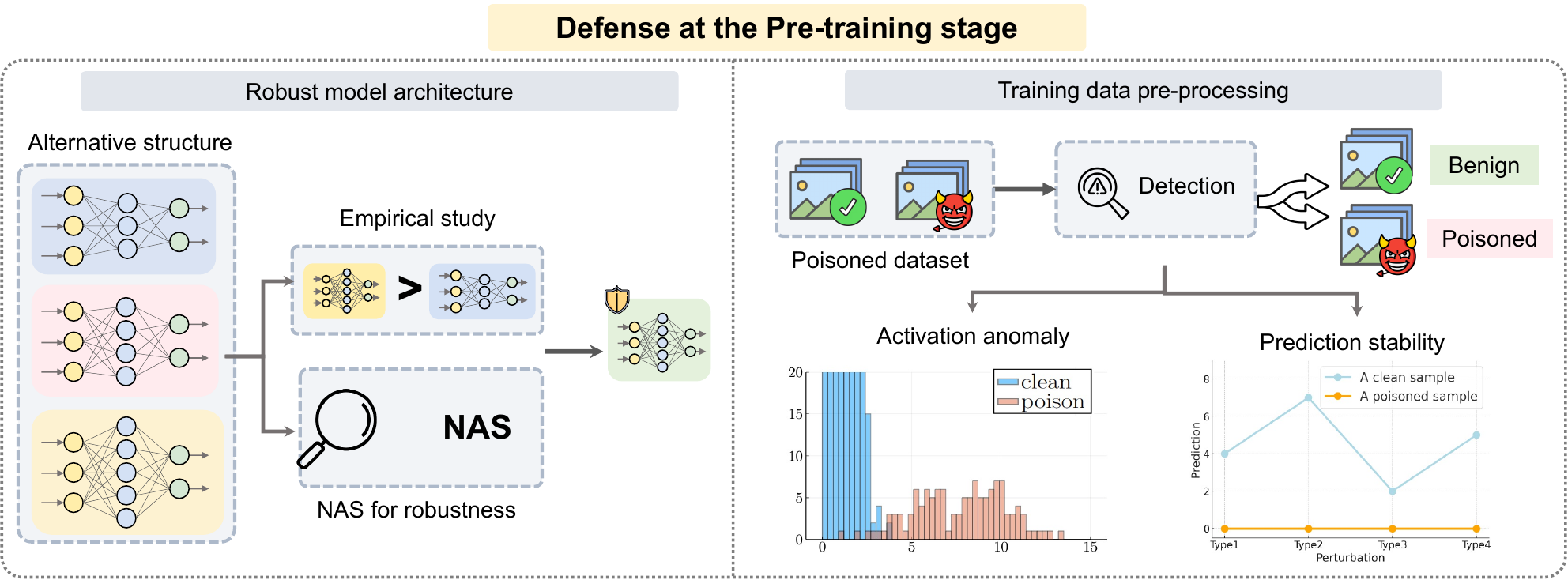}
    \caption{Graphical illustration of main defense strategies at the pre-training stage.}
    \label{fig:pretrain}
\end{figure*}

The pre-training stage serves as the initial phase for preparing the essential components required for subsequent training, encompassing the model architecture and training data. Consequently, defense mechanisms at the pre-training stage can be implemented by either designing a robust model architecture or pre-processing the training data, as shown in Figure~\ref{fig:pretrain}.

\subsection{Robust model architecture} 

It has been consistently observed in prior research that when provided with identical training data and training algorithms, models with distinct architectures often exhibit significant disparities in their levels of adversarial robustness. This observation has spurred a considerable body of work aimed at exploring whether certain architectures inherently possess greater adversarial robustness than others.
Two avenues have been pursued to identify robust architectures: one involves the empirical discovery of robust architectural features through experiments (\ie,  \textbf{Empirical study}), while the other employs Neural Architecture Search (NAS) algorithms to uncover unknown robust structures (\ie,  \textbf{NAS for robustness}).

\paragraph{Empirical study} Based on both theoretical research and experimental findings, two key factors significantly influence the robustness of architectural designs: 
\textbf{1) Model capacity.} Madry \etal \cite{madry2017towards} suggest that only models with capacities larger than traditional ones exhibit resilience against adversarial examples. Gao \etal \cite{gao_convergence_2019} expand this claim and establish a theorem elucidating the relationship between VC dimension and model robustness. Subsequent works such as \cite{wang2019convergence,xie_intriguing_2019} further substantiate these findings through experimental evidence, demonstrating that increased width and depth significantly enhance network robustness. Besides, some works \cite{huang_exploring_2021,peng_robarch_2023,wu_wider_2021,tang_robustart_2021,peng_robust_2023} delve into the intricate relationship between depth and width in model architectures and their impact on robustness. Specifically, Huang \etal\cite{huang_exploring_2021} propose that the narrowness of the final layer significantly influences model robustness, while WAR \cite{wu_wider_2021} emphasizes the need for adapting regularization terms according to the network depth to achieve heightened model robustness. Moreover, Peng \etal\cite{peng_robust_2023} put forward the notion of an optimal depth-to-width ratio for enhancing network robustness. In addition to investigating the model's intrinsic capacity, considerable attention has been devoted to exploring techniques for augmenting model parameters to bolster robustness. PSSiLU \cite{dai_parameterizing_2022} utilizes the implementation of parametric activation functions to further enhance model robustness. In \cite{qendro_towards_2022,zhou_bert_2020,hu_triple_2019}, the authors find that the adoption of multi-exit multiple classifiers not only accelerates inference speed but also elevates the model's representational ability, thereby bolstering its robustness. 
\textbf{2) Model structure.} Neural networks comprise various components, encompassing distinct activation functions, normalization layers, and connectivity patterns, which have different effects on model robustness. MBN \cite{xie_intriguing_2019} highlights the substantial impact of mean and variance in the Batch Normalization (BN) layer on adversarial robustness, and shows that the normalization of different types of samples is extremely different. Additionally, SAT \cite{xie_smooth_2020} provides evidence supporting the reduction of adversarial effects when employing smooth activation functions. Meanwhile, SGM \cite{wu_skip_2019} emphasizes the robustness-enhancing capabilities of skip connections within ResNet model. Furthermore, RobustMQ \cite{xiao_robustmq_2023} delves into the advantages of quantization models, showcasing their superior adversarial robustness compared to traditional models and a corresponding decrease in adversarial vulnerability as quantization bit width increases. FPCM \cite{bu2023towards} finds that the adversarial sample is more dependent on high-frequency information. Therefore, they design a module to reduce the high-frequency information in the learning process, which can improve the robustness of the model. As the application of Vision Transformers (ViTs) gains momentum, extensive attention has been devoted to examining their adversarial robustness in contrast to traditional Convolutional Neural Networks (CNNs). In \cite{shao_adversarial_2022,huang_exploring_2021}, the authors mention that ViTs exhibit superior adversarial robustness compared to traditional CNNs. However, they are more prone to interference from patches owing to their emphasis on low-frequency information and structural resemblance to randomly integrated models. In contrast, Bai \etal \cite{bai_are_2021} present an opposing perspective, supported by experiments demonstrating ViTs' resilience to patch perturbations. Furthermore, Liu \etal \cite{liu_exploring_2023} provide evidence of the disparity in adversarial complexity between ViTs and CNNs based on Rademacher complexity, showcasing how higher weight sparsity contributes to improving adversarial robustness in transformers. Building upon these insights, several works \cite{peng_robarch_2023,tang_robustart_2021} conduct an extensive series of experiments across various components and model architectures, culminating in the formulation of strategies for modifying model structures to enhance robustness.

\paragraph{NAS for robustness} 
The architectural structure itself plays a pivotal role in determining adversarial robustness, prompting the exploration of robust structures beyond conventional network designs. Guo \etal \cite{guo_when_2020} conduct a study demonstrating a strong correlation between architectural density and adversarial accuracy in the context of an adversarial robustness test based on Neural Architecture Search NAS. This research underscores the substantial enhancement in model robustness achieved through the adoption of dense connection modes. The pursuit of adversarially robust network structures via NAS encompasses two primary approaches: \textbf{1) Direct leveraging of adversarial robustness.} E2RNAS \cite{yue_effective_2022} introduces an efficient and effective Neural Network Architecture Search method, which directly incorporates adversarial robustness into the search process while considering performance and resource constraints. Another approach, RNAS \cite{zhu_robust_2023}, balances accuracy and robustness through the design of a regularization term, resulting in highly accurate and robust architectures. Moreover, G-RNA \cite{xie_adversarially_2023} employs NAS to discover robust graph neural networks. \textbf{2) Indirect utilization of robust network properties.} AdvRush \cite{mok_advrush_2021} identifies robust structures based on loss smoothness, while RACL \cite{dong_adversarially_2020} estimates the Lipschitz constant of the structure before conducting NAS to find the optimal configuration. Additionally, Dsrna \cite{hosseini_dsrna_2021} formulates two differentiable metrics, grounded in lower bound proofs and Jacobi norm bounds, to measure architectural robustness and subsequently seeks robust architectures by maximizing these metrics. To expedite NAS searches, ABanditNAS \cite{chen_anti-bandit_2020} streamlines the search space by estimating upper and lower bounds, enhancing search efficiency. DS-Net \cite{du_learning_2021} restricts the search space to a fixed block of architectural elements. Meanwhile, Wsr-NAS \cite{cheng_neural_2023} introduces a lightweight adversarial noise estimator to reduce the computational burden of generating adversarial noise with varying intensities. Additionally, they propose an efficient wide-spectrum searcher to minimize the cost associated with utilizing large adversarial validation sets for adjusting network structures during the search process. Lastly, CRoZe \cite{ha_generalizable_2023} adopts an innovative approach by concurrently applying adversarial learning and regular learning in parallel, facilitating information exchange through alignment and significantly accelerating the search process.

Note that the evaluations in all the above methods are conducted for robustness against inference-stage adversarial examples. In contrast, in the field of backdoor attacks, several work has been done for designing modules to defend against backdoor attacks. Tang \etal \cite{Tang2023SettingTT} find that backdoor attacks can be quickly learned by the network at the shallow layer, so they design a honeypot to limit the backdoor to the shallow layer of the network, to resist the backdoor attacks. We anticipate future research endeavors that will delve further into examining the efficacy of various structures in resisting backdoors.

\subsection{Training data pre-processing based defense against backdoor attack} 
\label{sec: pre-training defense against backdoor}

In this setting, the defender aims to identify the malicious samples, \ie, poisoned samples, and then discard them or apply specific operations such as purification or relabelling to clean up the dataset. Identification can be achieved by capturing the special behavior of poisoned samples, such as activation anomalies or prediction consistency. According to the behavior of poisoned samples utilized by the defender, the current pre-processing based defenses can be categorized into the following two types:

\paragraph{Activation anomaly in one backdoor model} The intuition behind these methods is that the activation (also known as embedding, feature or representation) in the latent space captures more compressed semantic information, which helps to discover the difference between poisoned samples and clean samples. For example, AC \cite{ac} reshapes each activation to a 1D vector and then reduces the dimension of all activations by Independent Component Analysis (ICA). Through performing \textit{k-means} with $k=2$ for activations in each class and computing the silhouette score for each cluster, they can separate the suspicious activations from clean activations, and identify the poisoned samples. Spectral \cite{spectral} utilizes singular value decomposition to obtain the top-$k$ right singular vector of the matrix of centered activations for each class. Then, they calculate outlier scores for samples and remove the top fraction of them, because poisoned samples will provide a strong signal for classiﬁcation. Different from Spectral, SPECTRE \cite{spectre} proposes $k-$IDENTIFIER algorithm to find the top-$k$ left singular vector of the activation matrix and the target label. Besides, it uses QUantum Entropy (QUE) scores to ﬁlter out samples with strong spectral signatures. To solve the problem that the features of poisoned samples can be deeply fused into those of clean ones, Tang \etal \cite{demon} propose SCAn which decomposes activation precisely. The detail of this method is that they leverage two-component decomposition on the activation of a subset of clean samples and estimate covariance matrices for each class. Based on these matrices, they calculate the identity vectors for the other samples and perform a likelihood ratio test and an EM algorithm to find the target class(es) and detect poisoned samples. Similarly to SCAn \cite{demon}, Beatrix \cite{beatrix} also needs some clean samples. In Beatrix, they use Gram matrix and its high-order forms to recover differences in activation between the poisoned samples and the clean samples. Unlike the methods mentioned above, NC(detection) \cite{wang2019neural} detects poisoned samples by examining whether the inputs activate poisoned neurons which are located by means in other stages discussed later. The sample-distinguishment module in D-ST \cite{chen2022effective} can identify poisoned samples by adding extra transformations to inputs and measuring their activation sensitivity to these transformations.

\paragraph{Prediction stability to model variation} Some detection methods detect poisoned samples by observing the output results (\eg, predictions or loss values), reducing the requirement of accessing intermediate activation. In most cases, poisoned samples are more sensitive to some operations, and have some abnormal behaviors. For example, ABL(detection) \cite{li2021anti} selects poisoned samples by filtering out the low-loss examples at an early training stage, which is one part of their defense method. CT \cite{qitowards} is a proactive poison detection approach that retrains a backdoor model on a confusion dataset consisting of clean samples with randomly mislabeled labels. After retraining, the model can only predict correctly on poisoned samples. Du \etal \cite{du2020robust} train a robust model on the contaminated dataset by introducing the noise satisfying the requirement of differential privacy. Then, the trained model performs similarly to the clean model and can distinguish the poisoned samples. ASSET \cite{asset} proposes a two-step optimization process to detect poisoned samples by training loss in various deep learning paradigms.

%% file: sections/section-training.tex
\section{Defense at the training stage}
\label{sec: defense at training}

\begin{figure*}
    \centering
    \includegraphics[width=1\linewidth]{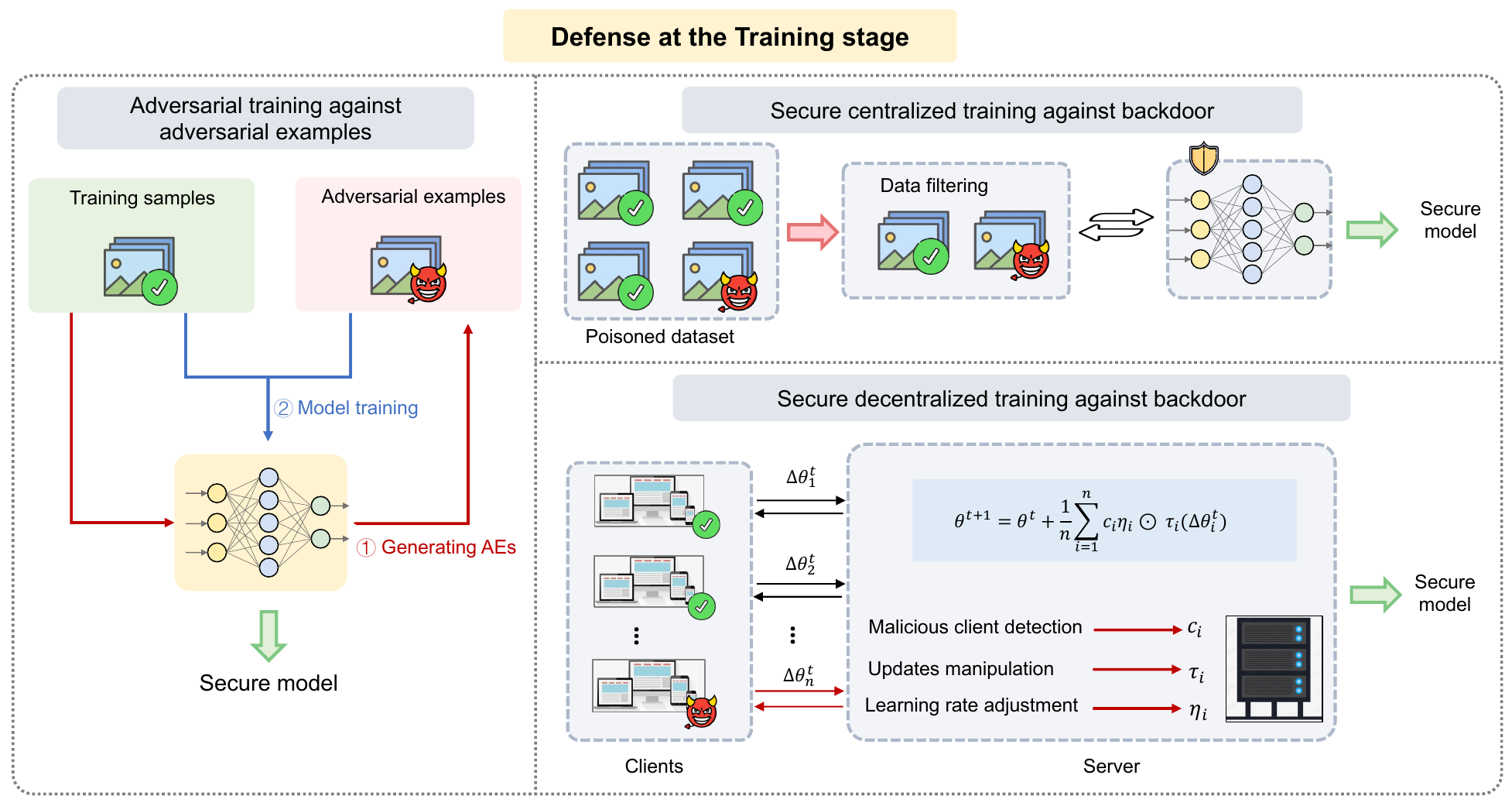}
    \caption{Graphical illustration of main defense strategies at the training stage.}
    \label{fig:train}
\end{figure*}

\subsection{Training-stage defense against  adversarial examples}

Adversarial training (AT) is the mainstream training-stage defense against adversarial examples by improving the model's robustness to adversarial examples. As shown in Figure~\ref{fig:train} (left), the general idea is to generate adversarial examples during the training process and make the model more resistant to them. The general formulation of AT is as follows:
\begin{equation}
    \min_{\boldsymbol{\theta}} \mathbb{E}_{(\x, y) \in \mathcal{D}} \left[ \mathcal{L}_1\big(f_{\boldsymbol{\theta}}(\x), y\big)  + \lambda \max_{\boldsymbol{\epsilon} \in \mathcal{S}} \mathcal{L}_2\big(f_{\boldsymbol{\theta}}(\x + \boldsymbol{\epsilon}), y\big) \right], 
\end{equation}
where $\mathcal{L}_1$ indicates the loss on clean data to guarantee clean accuracy, while $\mathcal{L}_2$ denotes the adversarial loss and $\lambda$ is a parameter to control the tradeoff between model robustness and accuracy. $\mathcal{S}$ represents a bounded space to restrict the bound of the perturbation $\boldsymbol{\epsilon}$. 

There are two main challenges of AT, \ie, \textbf{robust generalization} and \textbf{low efficiency}. The former refers to the poor adversarial robustness of the AT model to unseen adversarial examples that are not encountered during the training process. The latter is caused by the inner-max sub-problem, which makes the training cost of AT much higher than that of standard training. Existing AT works aim to tackle the above two challenges through various techniques. 

\paragraph{Enhancing the robust generalization}
To enhance the generalization ability of adversarial training, a line of research is to study the role of data and introduce various techniques to exploit 
the training samples. Adversarial Vertex mixup (AVmixup) \cite{lee2020adversarial} proposes a soft-labeled data augmentation approach that extends the training distribution by using linear interpolation. Cropshift \cite{li2023data}, along with a new data augmentation scheme, boosts robust generalization by improving the diversity and balancing the hardness of data augmentation. Diverse Augmentation based Joint Adversarial Training (DAJAT) \cite{addepalli2022efficient} improves robust generalization by using a combination of simple and complex augmentations with separate batch normalization layers in training. Besides, extra unlabeled and/or generated datasets have been shown to be effective for boosting robust generalization both empirically and theoretically \cite{alayrac2019labels, gowal2021improving, carmon2019unlabeled, xing2022artificially}.

Another perspective to train a model robust to various attacks, especially unseen attacks, is to improve the formulation of adversarial training. A simple idea is to adopt perturbation of multiple types \cite{liu2022mutual, dong2020adversarial, maini2020adversarial, tramer2019adversarial} 
or perturbation from different models \cite{yang2020dverge, tramer2017ensemble} in the training process. Besides, some work such as TRADES \cite{zhang2019theoretically} MART \cite{wang2020improving}  and SCORE \cite{pang2022robustness} focuses on the accuracy-robustness tradeoff and revises the objective of adversarial training to improve the robustness.
From the view of the weight loss landscape, a series of methods have been proposed to seek a flat solution to enhance the robust generalization. For example, Adversarial Weight Perturbation (AWP) \cite{wu2020adversarial} introduces adversarial perturbation to weights in the training process which explicitly regularizes the flatness of the weight loss landscape. Similar to AWP, other types of weight perturbation such as the robust weight perturbation \cite{ijcai2022p512} and the random weight perturbation \cite{jin2023randomized} to (critical) weight can also reduce the generalization gap. Recently, some methods have been proposed to measure the importance of samples for adversarial robustness and encourage models to focus on such important samples. Three types of Probability Margins (PMs)\cite{liu2021probabilistic} have been proposed to measure the AEs' closeness to decision boundary and reweight them based on the PMs. InfoAT \cite{xu2022infoat} employs the Information Bottleneck Principle and focuses on examples with high mutual information to enhance model robustness. Switching One-vs-the-Rest \cite{kanai2023one} (SOVR) adopts one-vs-the-rest loss for important samples, \ie, samples with small logit margins to achieve better robustness. Other measurements for sample importance including cross-entropy \cite{mendoncca2022adversarial} and robustness of samples \cite{yang2023improving}, have also been used for improving robust generalization.

Previous work observes that the strategy of attack also plays an important role in robust generalization. Based on such observation, some work has been done to improve robustness generalization by employing different strategies of attacks. For the initialization strategies, prior-guided FGSM initialization method \cite{jia2022boosting} and even learnable initialization \cite{jia2022prior} have been shown to be effective for generating high-quality AEs and therefore improving the model robustness. For the strength of attacks, Curriculum Adversarial Training \cite{cai2018curriculum} proposes to gradually increase the attack strength in the training process to enhance model robustness and Friendly Adversarial Training (FAT) \cite{zhang2020attacks} focuses on the least AEs minimizing the loss among all misclassified AEs. More recently, LAS-AT  \cite{jia2022adversarial} boosts adversarial training with a learnable attack strategy. Besides, Stabel-FGSM \cite{kim2021understanding} uses a simple method to choose a proper step size for FGSM. There are also some works aiming to improve robustness generalization by revisiting the optimization process \cite{dabouei2022revisiting}, introducing advanced optimization techniques such as the exponential moving average \cite{wang2022self}, or employing early stopping \cite{rice2020overfitting}.

\paragraph{Improving the training efficiency}
Adversarial training usually comes with high computation overhead, mainly due to solving the inner maximization problem.  To alleviate the computation cost of adversarial training, a simple strategy is to reduce the number of steps for constructing the AEs, such as FGSM with random start \cite{wong2020fast} which only takes one step for generating AEs with a random initialization. As FGSM-RS are known to suffer from catastrophic overfitting, an annealing mechanism (Amata) \cite{ye2021annealing} is proposed to dynamically alter the steps of attack. Another way to improve the efficiency of AT is to reduce the computation cost for each step. Free AT \cite{shafahi2019adversarial} reduces the cost of generating AEs by recycling gradient information in the updating of model parameters. You Only Propagate Once (YOPO) accelerates AT by focusing on the first layer of model \cite{zhang2019you}. Similarly, module robust criticality (MRC) is proposed in \cite{Zhu_2023_ICCV2} to measure the significance of specific module for robustness which can be fine-tuned to enhance the adversarial robustness efficiently.  Recently, some work has focused on selecting representative samples for adversarial training, \ie, only generating AEs for a portion of samples, therefore, improving the efficiency of AT. Coreset selection \cite{dolatabadi2022} has been shown to be effective in reducing the time complexity of adversarial training. Similarly, GRAD-MATCH and Adv-GLISTER \cite{li2023less} also prune the data to reduce the overhead of adversarial example generation.

\subsection{Training-stage defense against backdoor attack}

In this setting, the defender aims to train a model with high clean accuracy while simultaneously mitigating backdoor injection based on an untrustworthy training dataset.
According to the accessibility of the training dataset, existing training-stage defenses could be categorized into the following two types:

\paragraph{Secure centralized training against data-poisoning based backdoor attack} 
In this scenario, the defender has access to the entire training dataset but lacks knowledge about which specific samples are poisoned. Thus, most existing methods initially filter poisoned samples from the training dataset, and then a clean model could be trained based on the filtered training dataset as shown in Figure~\ref{fig:train} (right top), and sometimes, the poisoned samples are also incorporated into the training.
For example, anti-backdoor learning (ABL) \cite{li2021anti} identifies poisoned samples by observing training loss during the initial training epochs, based on the fact that the trigger in poisoned samples is relatively easy to learn, resulting in a rapid drop in training loss for these poisoned samples in the early epochs. Then, the filtered poisoned samples are unlearned to mitigate the model's backdoor effect in the later training process.
Decoupling-based backdoor defense (DBD) \cite{dbd} reveals that supervised learning leads to cluster poisoned samples in feature space, facilitating backdoor injection. Thus, it designs a three-stage training algorithm. Firstly, the backbone is trained via self-supervised learning \cite{simclr} based on label-removed training samples, to prevent the poisoned gathering. Then, on top of the frozen backbone, the linear classifier is trained via supervised learning, and the samples with high losses are identified as poisoned. Finally, the whole model is fine-tuned based on the training dataset by removing the poisoned samples' labels. 
In \cite{chen2022effective}, two defenses are presented based on the filtered training dataset according to the FCT metric (described in Section \ref{sec: pre-training defense against backdoor}). Distinguishment and secure training (D-ST) first trains the backbone via semi-supervised contrastive learning based on the dataset with poisoned labels removed from the dataset and then trains the linear classifier using mixed cross-entropy (MCE) loss that learns clean samples while unlearning poisoned samples. Distinguishment and backdoor removal (D-BR) is used to remove the backdoor from a pre-trained backdoored model by alternatively unlearning poisoned samples and relearning clean samples. 
Adaptively splitting dataset based defense (ASD) \cite{asd} 
dynamically splits the training dataset into clean data and potentially poisoned data pools, through loss-guided split and meta-learning-inspired split. In meta-split, clean samples are selected by the difference of loss between the current model and its twin model that is slightly fine-tuned on the whole training dataset.
An exception to the poison filtering strategy is the Causality-Inspired Backdoor Defense (CBD) \cite{cbd}. CBD firstly trains a backdoored model $f_B$ and then proceeds to train a clean model, $f_C$, by encouraging it to learn core label prediction information while being independent of the feature embedding present in $f_B.$

\paragraph{Secure decentralized training against partially training controllable backdoor attack}
In decentralized training, the training data and training process are distributed among several participants, and no participant can access the whole training dataset. To the best of our knowledge, most existing backdoor attacks against distributed training focused on federated learning \cite{bagdasaryan2020backdoor,xie2019dba,wang2020attack}. Therefore, we also focus on the backdoor defense in federated learning here, where the defender is located at the central server and cannot access any local training data, while the adversary operates on some clients and aims to compromise the model during the training process as shown in  Figure~\ref{fig:train} (right down).
The defender's goal is to stop the backdoor injection from unknown malicious clients during the training process, to obtain a secure global model.
As shown in the down-right corner of Figure \ref{fig:train}, to achieve that goal, there are three main technologies, including \textit{malicious client detection}, \textit{manipulating backdoor-related neuron updates}, and \textit{adjusting the learning rate of each neuron}.
The general formulation of both malicious client detection and manipulating backdoor-related neurons' updates is as follows:
\begin{equation}
    \Delta \boldsymbol{\theta} = \sum_{i=1}^n c_i \eta_i \odot \tau(\Delta \boldsymbol{\theta}_i), 
\end{equation}
where $c_i$, $\eta_i$, $\Delta \boldsymbol{\theta}_i$ correspond to (normalized) weights, learning rates, and updates of each client (from 1 to n), and $\tau$ represents any manipulation on updates of each client before aggregation.
\textbf{1) Malicious client detection} aims to identify malicious clients by utilizing some client-level discriminative characteristics and remove the malicious updates in the training process.
One commonly adopted discriminative characteristic is the client's model update/gradient.
Sniper \cite{cao2019understanding} observes that the malicious client's local model is far from that of a benign client. Prior to aggregation, a graph is constructed to represent the similarity among local models. Then, a maximal clique is identified from the graph, and the corresponding local models are aggregated to update the global model. Alternatively, the weight ($c_i$) of non-selected clients is set to 0.
As for Li \etal \cite{li2020learning}, an autoencoder is trained based on updates from all clients' local models, and clients exhibiting a high reconstruction error are identified as potential malicious entities.
FLAME \cite{nguyen2022flame} utilizes a dynamic clustering algorithm called HDBSCAN \cite{hdbscan} on all local model updates, and those with high angular deviations from the majority of updates are identified as malicious clients.
The defending method PA-SM \cite{lu2022defense} against convergence-round backdoor attack (\ie, initiated at convergence stage) assumes that if the cosine similarity between one local model update ($\Delta \boldsymbol{\theta}_i$) and the aggregated model update ($\Delta \boldsymbol{\theta}$) is close to 1, then it is identified as malicious.
The defending method ACCDR \cite{lu2022defense} against early-round backdoor attack (\ie, initiated at early stage) firstly recovers a trigger through optimization, then measures the activation difference between a random noise with and without trigger on each local model. The local model with the large difference is considered malicious. 
Recently, FedCPA\cite{Han_2023_ICCV} utilizes critical parameter analysis, since they observe benign model updates share more similar top and bottom important parameters than malicious model updates.
As for Huang \etal \cite{Huang_2023_ICCV}, multiple metrics are considered at the same time to deal with diverse malicious gradients that cannot be well addressed by single metric detection methods.
The defending method Fed-FA\cite{fedFa_2023} observes the deficiency of present robust federated aggregation on NLP tasks due to discrete feature space, so they identify backdoor clients through modeling data divergence among clients with approximation of clients’ data Hessian. 
FedGame\cite{FedGame_2023} address the limitation of present defense methods that they focus on static attacker model and inadequate attention is paid to dynamic attackers who adapt their attack strategies dynamically. They reverse engineer backdoor trigger and target class to compute a defined genuine score for each client, which denotes the ratio of trigger-embedded input being classified to target class. The weighted sum utilizing genuine score can then be robust to backdoors.
\textbf{2) Manipulating backdoor-related neuron updates} aims to perturb or modify the updates of neurons that are possibly related to backdoor effect, to stop the backdoor injection.
For example, in addition to detecting malicious clients, FLAME \cite{nguyen2022flame} also designs an adaptive clipping depending on the $\ell_2$-norm of model update, and an adaptive noise that depends on the sensitivity and $\ell_2$-norm of the local model weights.
As for Sun \etal \cite{sun2019can}, a norm-clipping approach through normalizing the model updates and a small random noise on each model update are adopted together to mitigate the backdoor effect.
Robust filtering of one-dimensional outliers (RFOut-1d) \cite{rodriguez2022backdoor} assumes that for each neuron, if its update is far from the mean update across all participated clients in aggregation, then it is likely to be backdoor-related, and its update is replaced by the mean update to prevent backdoor injection.
PartFedAvg \cite{gao2022secure} randomly chooses partial parameters from each client's model update in aggregation, \ie, randomly setting some parameters in each model update to zero, to mitigate the potential backdoor effect.
\textbf{3) Adjusting the learning rate of each neuron} aims to set small or even negative learning rate for the backdoor-related neuron, while larger rates for other neurons. For example, FoolsGold \cite{fung2018mitigating} assumes that the client-level aggregated historical gradients of two malicious clients ($\Delta \boldsymbol{\theta}$) exhibit a higher degree of similarity. This observation leads to the adjustment of each client's learning rate ($\eta_i$) based on its maximal cosine similarity with all other clients.
Robust learning rate (RLR) \cite{ozdayi2021defending} adjusts the learning rate for each neuron when updating the global model. Specifically, if the sum of gradient signs \wrt  one neuron across all local clients is larger than some threshold, \ie, the update directions for that neuron are consistent, then a positive learning rate is adopted, and \emph{vice versa}.

%% file: sections/section-post-processing.tex
\section{Defense at the post-training/deployment stage}
\label{sec: defense at post-training}

\begin{figure*}
    \centering
    \includegraphics[width=1\linewidth]{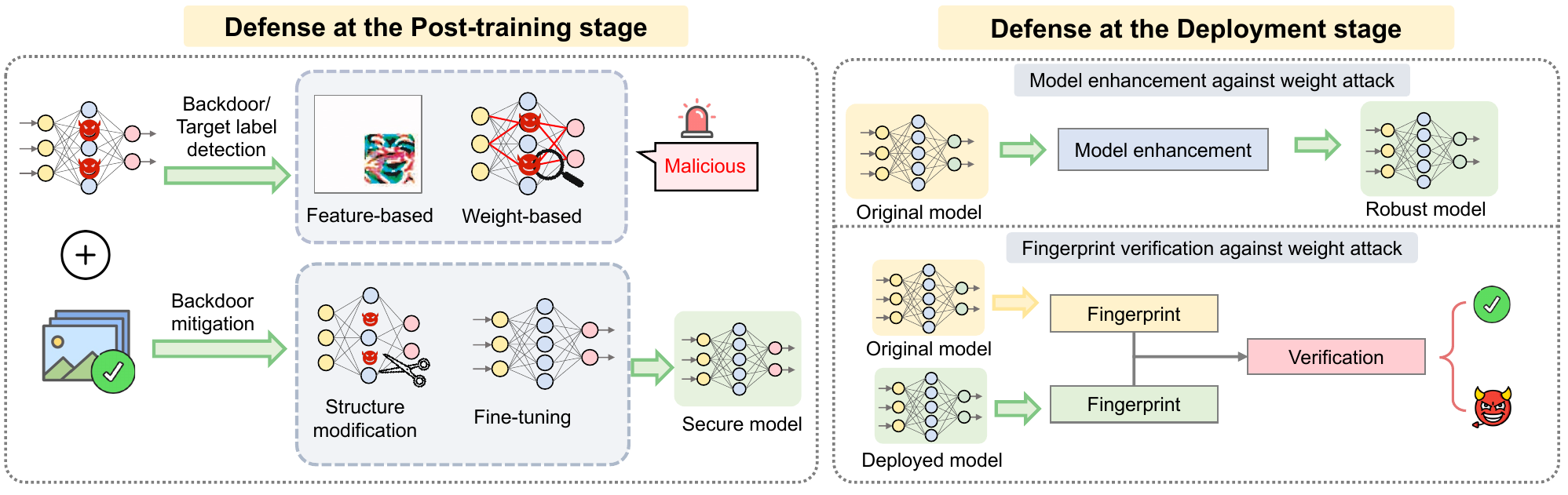}
    \caption{Graphical illustration of main defense strategies at the post-training (left) and deployment (right) stage, respectively.}
    \label{fig:posttrain}
\end{figure*}

\subsection{Post-training-stage defense against backdoor attack}
In this setting, given an untrustworthy model that may contain backdoor, the defender aims to obtain a secure model by removing the potential backdoor while preserving the clean performance.
As shown in Figure~\ref{fig:posttrain} (left), this goal could be achieved by three main tasks, including \textbf{backdoor detection}, \ie, detecting whether the given model is backdoored or not; \textbf{target label prediction}, \ie, detecting the target label(s); \textbf{backdoor removal/mitigation}, \ie, modifying the backdoored model to remove the possible backdoor or mitigate the backdoor effect without significantly harming the model's clean accuracy. It is commonly assumed that the defender can access a few clean samples in this paradigm.

\paragraph{Backdoor detection} 
Existing backdoor detection methods could be categorized into two types. \textbf{1) Feature-based detection} methods detect backdoors by inspecting some abnormal phenomena in the feature space of the model, especially on potential poisoned samples. However, the main challenge is the lack of prior information on the potential trigger and target class. Thus, several methods attempt to first reverse or approximate the potential trigger. Then the distinctive characteristics of potential poisoned samples are utilized to detect backdoors.
ABS \cite{liu2019abs} first identifies potentially compromised neurons by connecting them with the activation elevation of the output label. Then it employs reverse engineering techniques to confirm these compromised neurons, and a model is backdoored if the trigger can subvert all the benign inputs to the same output label.
Hu \etal \cite{hu2022trigger} aim to create precise, high-quality triggers by leveraging innovative priors such as diversity and topological simplicity. These diverse trigger candidates are subsequently employed to train a shallow neural network for backdoor detection.
EX-RAY \cite{liu2022complex} first searches universal input patterns to reverse triggers. Then, symmetric feature differencing calculates masks that reveal discrepancies between a clean sample and its poisoned counterpart, as well as between the poisoned instance and a sample from the target class. These dissimilar masks serve as indicators for detecting backdoored models.
DECREE \cite{feng2023detecting} is the first backdoor detection approach for pre-trained encoders. It seeks minimal trigger patterns, ensuring that inputs marked with the trigger exhibit similar embeddings. A large $L^n$ norm of the identified trigger is likely a backdoored encoder.
DeepInspect \cite{chen2019deepinspect} addresses a challenging scenario involving a black-box detection model and inaccessible to clean samples. It starts by recovering a substituted training dataset through model inversion, followed by learning the trigger distribution using a generative model, and subsequently employs statistical hypothesis testing for anomaly detection.
One-Pixel Signature \cite{huang2020one} detects backdoors by introducing the one-pixel signature of a network, which is the collection of single-pixel adversarial perturbations that most effectively impact the label of a collection of images.
Zheng \etal \cite{zheng2021topological} use topological tools to model high-order dependencies in the networks and use these topological features to comprehensively evaluate the existence of backdoors.
\textbf{2) Weight-based} methods utilize the weights of a detected model as input features and subsequently train a meta-classifier designed specifically for the detection of backdoors. 
ULPs \cite{kolouri2020universal} first trains hundreds of clean and poisoned models. Then a binary detection classifier along with learnable Universal Litmus Patterns (ULPs) is trained using the output of these hundreds of models. The detection of backdoors is achieved by feeding the ULPs to the identified models.
MNTD \cite{xu2021detecting} trains a meta-classifier to detect backdoors which first trains a number of shadow models, and these models serve as input to train the meta-classifier.

\paragraph{Target label prediction} 
This task aims to not only detect backdoors, but also the target label(s) of attackers. Existing target label prediction methods can be divided into feature-based approaches, which detect backdoors by inspecting abnormal phenomena in feature space; and weight-based approaches, which analyze weight characteristics to detect backdoors.
\textbf{1) Feature-based approaches.}
NC \cite{wang2019neural} first reverses a masked trigger by minimal universal perturbation for each label and then detects target label by outlier detection.
TND \cite{wang2020practical} considers both data-limited and data-free scenarios. It detects target labels by comparing the similarity between adversarial examples under the universal perturbation and image-wise perturbation for each class and images are replaced by random noise in data-free cases.
Since the problem complexity is quadratic to the number of class labels in many detection methods, K-Arm \cite{shen2021backdoor} proposes an efficient detection method that utilizes K-arm strategy borrowed from reinforcement learning to efficiently detect target labels. 
Considering data-scarce cases, L-RED \cite{xiang2021red} proposes an efficient Lagrangian-based reverse-engineering-based detection method utilizing pattern estimation and anomaly detection.
Dong \etal \cite{dong2021black} propose a black-box backdoor detection (B3D) method which designs a gradient-free optimization algorithm to reverse the trigger for each class and detect target labels based on the small trigger norms.
Guo \etal \cite{guo2022aeva} propose the adversarial extreme value analysis (AEVA) to detect target labels in black-box models, which includes estimating gradient by zeroth-order estimation and detecting target labels based on an extreme value analysis of the adversarial map.
Recently, Xiang \etal \cite{xiang2022post} address multiple attacks problem and introduce a novel expected transferability (ET) based on reverse engineering to detect target labels. In short, ET is the probability of adversarial perturbation for one sample is also applicable to others within the same class.
\textbf{2) Weight-based approaches.}
Greg \etal \cite{fields2021trojan} focus on the final linear layer of the network and associate trojan target label with a larger average weight of the target row in the linear layer. 
CPBD \cite{jiang2022critical} proposes a critical-path-based backdoor detector that formulates a set of critical paths for a neural network and detects the target labels by analyzing critical paths in each class.

\paragraph{Backdoor removal/mitigation}
Even knowing that there is a backdoored model, it is still challenging to identify where the backdoor is, as well as remove or mitigate the backdoor from the model. There are two general approaches.   
\textbf{1) Structure modification approaches} can be divided into pruning-based methods, which aim to identify which neurons contribute to backdoor and prune them; and augmenting model parameters methods, which aim to filter out or suppress backdoor-related features by inserting additional parameters. 
FP \cite{liu2018fine} prunes the neurons that cannot be activated by benign data and then fine-tuning the pruned model to improve model utility. 
ANP \cite{wu2021adversarial} finds that neurons associated with backdoors exhibit heightened susceptibility to adversarial perturbations. It employs a minimax optimization to identify and mask these compromised neurons. 
Considering a data-insufficient situation, 
ShapPruning \cite{guan2022few} considers the interaction among neurons and detects compromised neurons through the computation of their Shapley values. It subsequently mitigates the backdoor by pruning the neurons with the highest Shapley values.
Purifier \cite{zhang2022purifier} observes backdoor triggers consistently manifest similar anomaly activation patterns in intermediate activation and introduces a Plug-and-play module to suppress these anomaly activation patterns. 
To further improve pruning efficiency, CLP \cite{zheng2022data} proposes a data-free pruning strategy that measures potential backdoor channels by the proposed Channel Lipschitz Constant. 
Zheng \etal \cite{zheng2022pre} observe notable distinctions in the moments of pre-activation distributions between benign and poisoned data within backdoor neurons, as opposed to clean neurons. Building upon this observation, they introduce two pruning strategies, namely EP and BNP.  
AWM \cite{chai2022one} is also a data-efficient approach that employs minimax optimization to recover trigger patterns in an adversarial manner and applies (soft) weight masking to the model's sensitive neurons influenced by these patterns.
NPD \cite{zhu2023neural} introduces a poisoned feature purification method that involves the insertion of a lightweight linear transformation layer into the poisoned model. It is efficient since only the linear transformation layer is learned via bi-level optimization without altering the original model's parameters.
Some methods also investigate backdoor mechanisms from model structure's perspective. For instance, SSFT \cite{yang2023backdoor} posits a strong correlation between skip connections and backdoors, and it mitigates backdoors by suppressing skip connections in crucial layers.
\textbf{2) Tuning-based approaches without structural modification} aim to mitigate the backdoor effect by tuning model parameters with some clean training data. Tuning-based approaches can be divided into data-based methods, which mitigate backdoors with the help of data augmentation or data generation; and objective-based methods, which formulate different objective functions to mitigate backdoors. For data-based method, 
DeepInspect \cite{chen2019deepinspect} is the pioneering black-box backdoor defense method. It recovers a substituted training dataset through model inversion, learns the trigger distribution with a generative model, and employs the trigger generator for effective backdoor mitigation.
For objective-based methods, a standard formulation is:
\begin{equation}
    \min_{\boldsymbol{\theta},m,\blacktriangle}\mathbb{E}_{(\x, y) \in \mathcal{D}}\mathcal{L}(f(A(\x,m,\blacktriangle),{\boldsymbol{\theta}}),y), 
\end{equation}
where $\blacktriangle$ and $m$ denote the reversed trigger and mask on the trigger, respectively. $A$ denotes the fusion function for poisoned samples. 
NC \cite{wang2019neural} and i-BAU \cite{zeng2022adversarial} first reverse triggers with minimal universal perturbations and then relearn these generated substituted backdoor samples to mitigate backdoors. 
Compared to treating the trigger as a single point, MESA \cite{qiao2019defending} approximates the trigger as a distribution by max-entropy staircase approximator for generative modeling. 
Recently, PBE \cite{mu2023progressive} points out that untargeted adversarial samples and their triggered images can activate the same DNN neurons. Therefore relearning on adversarial samples can purify the infected model.
FT-SAM \cite{Zhu_2023_ICCV} proposes a new defense method which searches for a new minima based on adaptive sharpness-aware minimization.
Recently, Pang \etal \cite{pang2023backdoor} question the requirement of training data. They propose to use unlabeled samples to purify models, which adopts an adaptive layer-wise initialization and knowledge distillation.
FST \cite{min2023towards} shows that fine-tuning methods fail at low poisoning ratios since the entanglement between backdoor and clean features undermines the effect of tuning-based defenses.
They design a stronger tuning method to shift backdoor features by encouraging the change of linear classifiers.
SAU \cite{wei2023shared} establishes the relationship between adversarial examples and poisoned samples and derives a novel upper bound for the backdoor risk, with which they propose a bi-level optimization problem for mitigating the backdoor using adversarial training techniques.
There are also some works devoted to purely trigger reversion engineering, such as UNICORN \cite{wang2023unicorn}, SmoothInv \cite{sun2023single}, and BTI \cite{tao2022better}.
Apart from trigger inversion methods, there are some defense methods drawing support from additional models. 
For example, Zhao \etal \cite{zhao2020bridging} shows that path connections trained by a small proportion of clean samples can significantly purify backdoored models.
NAD \cite{li2021neural} firstly fine-tunes a teacher model with clean samples and the teacher model can foster a better backdoor removal for the student model.

\subsection{Deployment-stage defense against weight attack}
\label{sec: defense at deployment}

While the adversarial weight attacks occur at the deployment stage, there are also some defenses adopted at the same stage to mitigate the weight attack effect. Existing methods can be divided into model enhancement based methods, which enhance the model's resistance to weight attacks, and fingerprint verification based methods, which generate the signature of model and check the signature to detect the weight attack.
\textbf{1) Model enhancement based.} The output code matching (OCM) method \cite{ozdenizci2022improving} proposes to replace the usual one-hot encoding of classes by partially overlapping bit strings, through changing the softmax layer to a N-length output layer with tanh activation function, such that the bits of representing different classes are correlated.
Consequently, the cost to achieve a successful weight attack to a particular target class is increased, and the benign accuracy of other classes is significantly influenced. 
The Aegis method \cite{wang2023aegis} observes that existing targeted weight attacks often flipped some critical bits in higher layers (\ie, layers close to the output). Inspired by this observation, Aegis adopts a dynamic-exit mechanism to attach extra internal classifiers (ICs) to hidden layers, such that the inference of each input sample could early-exit from different intermediate layers, to mitigate the weight attack effect in higher layers. 
The randomized rotated and nonlinear encoding (RREC) method \cite{liu2023generating} adopts two strategies to enhance the tolerance to bit flips in weight attack, including randomized rotation to obfuscate the bit order of weights, to reduce the chance of identifying vulnerable bits, and nonlinear quantization function of weights to reduce the bit-flip distance. 
\textbf{2) Fingerprint verification based.} 
The DeepAttest method \cite{chen2019deepattest} firstly designs a device-specific fingerprint, and encodes it into the model weights deployed on the device, while storing it in a trusted execution environment (TEE) of the same device. Then, at the inference stage, the fingerprint is extracted from the model. If the extracted fingerprint is consistent with that stored in TEE, the model is verified and could be used for inference. 

In addition to the above defense methods designed for adversarial weight attacks (\ie, targeted), there are also several weight manipulation detection methods for any weight manipulation attacks, including both targeted and untargeted weight attacks. We refer readers to \cite{qian2023survey} for more details about these detection-based defense methods.

%% file: sections/section-inference.tex
\section{Defense at the inference stage}
\label{sec: defense at inference}
\begin{figure*}
    \centering
    \includegraphics[width=0.9\linewidth]{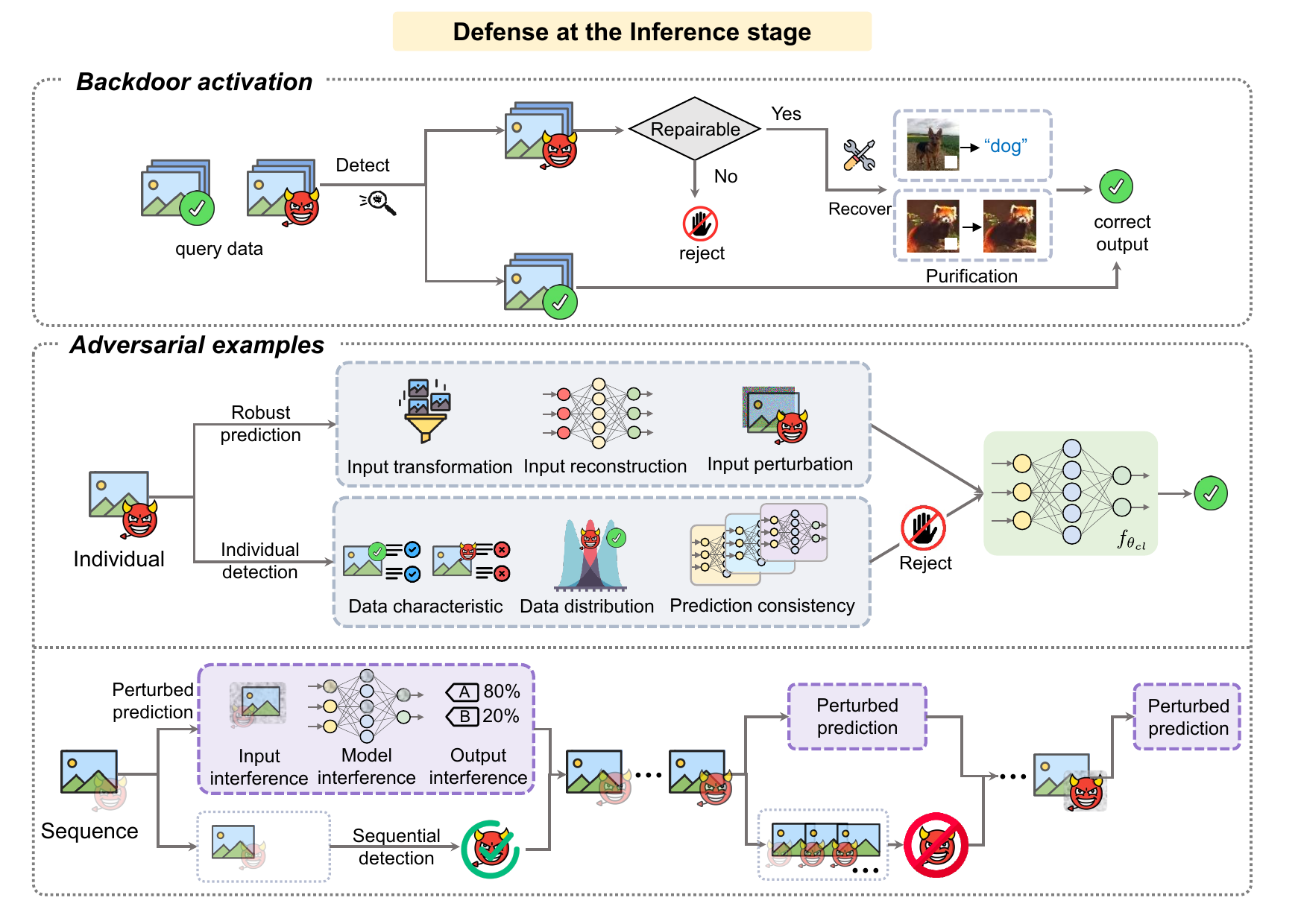}
    \caption{Graphical illustration of main defense strategies at the inference stage, against backdoor activation (top) and adversarial example attack (bottom), respectively.}
    \label{fig:inference}
\end{figure*}

\subsection{Inference-time defense against backdoor attack}
In this situation, the defender lacks information about the potential backdoor injection in the model. The defender's objective is to prevent the potential backdoor from being activated by queries with a particular trigger while keeping the model fixed. There are mainly two tasks in this stage, including poisoned sample detection and poisoned sample recovery, in which the model outputs the original label and original semantic information of poisoned query data, respectively.

\paragraph{Poisoned sample detection} Two main approaches have been developed to achieve this objective, including query data perturbation, and query data distinction.

\textbf{1) Query data perturbation}: STRIP \cite{gao2019strip} perturbs query inputs by superimposing various image patterns and calculating the entropy of predictions, where clean samples exhibit high entropy and poisoned ones typically yield lower entropy values. SentiNet \cite{sentinet} uses GradCAM to localize the salient region which could capture the suspected regions for classification. Then, this suspected region is overlaid on a set of benign samples, and determines whether this region can hijack the expected predictions of these benign samples. TeCo \cite{liu2023detecting} perturbs the inputs with various image corruptions and measures the corruption robustness consistency. They discover that the tendency of predictions under these corruptions is similar for clean samples, while it is disordered for the predictions of poisoned samples. SCALE-UP \cite{guo2023scale} also measures the prediction consistency by perturbing test samples with the pixel-wise ampliﬁcation process. CBD \cite{xian2023unified} uses the different distribution of the transformed backdoor samples and the clean samples to construct hypothesis testing statistics to judge the backdoor samples.
\textbf{2) Query data distinction}: When the defender possesses some prior understanding of poisoned or clean samples' characteristics, they can utilize these characteristics to detect poisoned samples in the inference stage. FreqDetector \cite{zeng2021rethinking} indicates that poisoned samples usually exhibit high-frequency artifacts, so a frequency detector is trained on a synthetic dataset for frequency based backdoor detection. FREAK \cite{freak} is another frequency-based detection method, which finds that the frequency sensitivity is different between poisoned and clean samples. They analyze the indices of top-$k$ sensitive frequency components of given clean samples and measure the distance between the indices of query inputs and those of clean samples to detect the outliers. Orion \cite{orion} attaches many small side networks to the multiple branches of the original backdoor model, which can be seen as a multi-exit branchy network. There are some given clean samples to train this multi-exit network. When the network can capture deviations between the poisoned and clean samples from shallow layers to deep layers, it can be used to identify poisoned samples in the inference stage.

\paragraph{Poisoned sample recovering} To purify poisoned samples, two main approaches are proposed, including poisoned sample relabeling and poisoned sample purification. \textbf{1) Poisoned sample relabeling:} Orion \cite{orion} discovers that shallow layers of networks usually learn the original features of samples, so they utilize these features to induce the possible original labels of poisoned samples. NAB \cite{nab} proposes two relabeling methods. Specifically, the defender can relabel poisoned samples by the model that is trained on a small clean validation dataset, or on the dataset that has removed some suspicious samples. \textbf{2) Poisoned sample purification:} ZIP \cite{shi2023black} leverages a linear transformation to destroy trigger patterns added to the inputs and utilizes a pre-trained diffusion model to complete semantic information. Once the semantic information is accurately filled in, the trigger becomes ineffective, thereby inhibiting the activation of backdoor. 

\subsection{Inference-time defense against adversarial examples}
During the inference stage, the model is expected to deliver high-quality responses to benign queries. Moreover, it should also be robust with potential threats, including \textbf{individual adversarial inputs} and \textbf{sequences of malicious queries}.

\paragraph{Individual adversarial inputs} To defend against individual adversarial inputs while keeping the model unchanged, defense strategies can be divided into giving correct feedback and rejecting the input as follows:

\textbf{1) Giving correct feedback.} By denoising or purifying the adversarial samples, the processed samples are fed into models to give correct feedback, termed \textbf{robust prediction}.
There are three ways to eliminate adversarial perturbation and give correct predictions. \textbf{\ding{182} Input transformation based.} Many works aim at removing adversarial perturbations from the input by applying simple and effective image processing steps before feeding them to the model. For the given model $f$, the general formulation of input transformation based robust prediction on the potentially attacked input $x_{test}$ is specified as:
\begin{equation}
 \hat{y}=f\left(T\left(x_{test}\right)\right),
\end{equation}
where $T(\cdot)$ is an image transformation function. Guo \etal \cite{guocountering} first demonstrate the effectiveness of five standard image transformations in defense: image cropping and rescaling, bit-depth reduction, JPEG compression, total variance minimization, and image quilting. Das \etal \cite{das2017keeping} and Feature Distillation \cite{liu2019feature} further counter attacks using direct-but-ensembled and DNN-oriented JPEG compression, respectively. Similar to JEPG compression, which aims to suppress high-frequency components (HFC), DAD (Data-free Adversarial Defense) \cite{nayak2022dad} disregards HFC while maintaining high discriminability. Song \etal \cite{song2018defense} utilize the Saak transform to compute spatial-spectral representations and filter out HFC to eliminate perturbations. Mustafa \etal \cite{mustafa2019image} employ super-resolution techniques to remove high-frequency adversarial noise. C. P\'erez \etal \cite{perez2021enhancing} conduct a comprehensive empirical study to investigate the effects of Test-time Transformation Ensembling (TTE) on adversarial defense. 
\textbf{\ding{183} Input reconstruction based.} Input reconstruction based methods within a distinct class involve acquiring a reconstruction model designed to restore the benign counterpart of an adversarially perturbed image before classification. Simultaneously, the reconstruction process must ensure that the classification outcomes for unaltered examples remain unaffected. If the reconstruction model $R$ is parameterized by $\theta$, then $R$ can be trained by minimizing:
\begin{equation}
\hat{\theta}=\arg \min_{\theta} \mathcal{L}_R\left(x_{\text {train }}, x_{\text {train }}^* ; \theta\right),    
\end{equation}
where $\mathcal{L}$ is the reconstruction loss specific to each method. Clean examples $x_{\text {train }}$ and, occasionally, crafted adversarial examples $x_{\text {train }}^*$ are used during training.
At inference time, the general formulation of input reconstruction based robust prediction is specified as:
\begin{equation}
\hat{y}=f\left(R\left(x_{\text {test }} ; \hat{\theta}\right)\right).
\end{equation}
One main idea of this category involves constructing a generative model to model the benign data distribution. With this model, the defender can guide the perturbed examples back to the benign distribution while preserving the original image features needed for classification.
MagNet \cite{meng2017magnet} introduces an autoencoder-based reformer. With training on the benign dataset, this autoencoder produces an example approximating the adversarial input yet closer to the benign manifold. Two defenses closely related to MagNet that implicitly approximate the distribution are Defense-GAN \cite{samangoueidefense}, which employs generative adversarial networks (GANs), and ComDefend \cite{jia2019comdefend}, which utilizes a compression model.
Hill \etal \cite{hillstochastic} conduct a long-run Langevin update on the input using a convergent energy-based model (EBM) trained on unlabeled benign examples. The memoryless long-run sampling erases adversarial signals while ensuring the preservation of image classes through convergence. Two closely related defenses to \cite{hillstochastic} that explicitly approximate the distribution are presented in \cite{yoon2021adversarial}, which employs denoising score-matching (DSM), and in DiffPure \cite{nie2022diffusion}, which utilizes a diffusion model. 
Another line of this category involves crafting adversarial examples to train a denoising layer/model by minimizing the difference between the clean example and its adversarial counterpart. During inference, the given base classifier is augmented with the custom-trained denoiser. Approaches following this line primarily focus on introducing various denoiser architectures under effective optimization guidance. 
High-level representation guided denoiser (HGD) \cite{liao2018defense} proposes a denoising U-net and optimizes it by minimizing the difference between high-level features of the adversarial image and the denoised image. Adversarial perturbation elimination with GAN (APE-GAN) \cite{jin2019ape} replaces the denoising autoencoder in \cite{liao2018defense} with GAN, training its denoiser (generator) to deceive the discriminator capable of distinguishing denoised examples from original clean images. Borkar \etal \cite{borkar2020defending} insert trainable feature regeneration units into the given model, rendering the most vulnerable filter activations resilient. Differential defense with local implicit functions (DISCO) \cite{hodisco} predicts the clean RGB values of input pixels by implementing an encoder and a local implicit function. Divide, denoise, and defend method (D3) \cite{moosavi2018divide} purifies the input with dictionary learning and sparse reconstruction, whose intuition is to improve robustness via reducing dimensionality. 
Unlike previous approaches aimed at eliminating specific adversarial noises, the denoiser of denoised smoothing (DS) \cite{salman2020denoised} removes the Gaussian noise added to clean images, providing certified robustness guarantees, and the denoiser of neural representation purifier (NRP) \cite{naseer2020self} removes the highly transferable perturbations, enabling a generalizable defense against unseen attacks.
\textbf{\ding{184} Input perturbation based.} Some attempts focus on modeling the intrinsic behaviors inherent in clean images. By doing so, the adversarial input could be perturbed back to its original clean state toward imitating the clean behaviors. Unlike methods based on input reconstruction, which seek a universal denoiser for all adversarial examples, methods based on input perturbation tailor the defensive perturbation for each adversarial instance. Given the method-specific perturbation loss $\mathcal{L}_P$, the general formulation of input perturbation-based robust prediction is specified as follows:
\begin{equation}
\begin{aligned}
& \hat{y}=f\left(x_{\text {test }}+\hat{\gamma}\right), \\
& \text {s.t. } \hat{\gamma}=\arg \min_{\gamma} \mathcal{L}_P\left(x_{\text {test }}+\gamma ; \hat{\theta}\right),
\end{aligned}
\end{equation}
where $\hat{\theta}$ is the parameter of the auxiliary model that is trained by minimizing:
\begin{equation}
\hat{\theta}=\arg \min_{\theta} \mathcal{L}_P(x_{\text{train }}, x_{\text {train }}^* ; \theta).
\end{equation}
PixelDefend \cite{songpixeldefend} hypothesizes that clean examples reside in high-density regions. Therefore, it estimates the clean distribution using PixelCNN and purifies the adversarial image by maximizing its likelihood. Hilbert-based PixelDefend (HPD) \cite{bai2019hilbert} enhances PixelDefend by substituting PixelCNN with Hilbert-based PixelCNN to further model pixel dependencies. Self-supervised online adversarial purification (SOAP) \cite{shionline} posits that clean examples excelling in self-supervised tasks should also excel in classification tasks. Hence, inputs are optimized, aiming for high performance in auxiliary self-supervised tasks. Self-supervised reverse attack \cite{mao2021adversarial} also leverages self-supervision but instead focuses on restoring the natural structure within the contrastive objective. Hedge defense \cite{wu2021attacking} and anti-adversaries \cite{alfarra2022combating} both attempt to shift adversarial examples away from the decision boundary towards regions where clean examples reside.

\textbf{2) Rejecting the input.} 
This type of defense adopts a strategy known as \textbf{adversarial individual detection}, wherein it rejects the input if any single input is identified as adversarial.
Although adversarial inputs stem from tiny, sometimes visually imperceptible, perturbations of benign examples, numerous studies have shown that they inherently possess a different distribution from benign examples. Thus, these studies successfully train a detector network that obtains intermediate feature representations of a classifier as distinguishable features to detect adversarial inputs. Metzen \etal \cite{metzendetecting} trains an auxiliary binary classification model on features extracted from the given model. Due to the vulnerability of \cite{metzendetecting} to Type II attacks, Safetynet\cite{lu2017safetynet} focuses on discrete features produced by quantizing activation in late-stage ReLUs to conceal the gradient. Grosse \etal \cite{grosse2017statistical} augment the given model’s output by introducing an extra class and training the model to identify adversarial examples within this new class.
DAD (data-free adversarial defense)\cite{nayak2022dad} extends to the data-free adversarial defense setting. It initializes the detection layer through training on arbitrary data and finetunes it using unlabeled test-time input.
Beyond the straightforward features extracted from the intermediate layer of the given model, more efforts by domain experts involve the intricate design of handcrafted feature extractors. As adversarial examples heavily impose a regularization effect on nearly all informative directions, Li \etal \cite{li2017adversarial} extract PCA statistics from convolutional filter outputs, constituting evident features, and devise a cascade classifier. 
Ma \etal \cite{macharacterizing} characterize and distinguish the adversarial region by local intrinsic dimension (LID), which is higher than that for the benign data region. Demonstrating that the Fisher information matrix (FIM) aptly captures local robustness, Zhao \etal \cite{zhao2019adversarial} employs the eigenvalues of the FIM as features for adversarial detection.
Cohen \etal \cite{cohen2020detecting} argue that any inputs deviating from the correlated pattern between the most influential training samples (measured by the influence function) and predictions (relying on nearest neighbors) are deemed adversarial.
Observing higher feature disagreement among adversarial examples, multi-layer leave-one-out (ML-LOO) \cite{yang2020ml} employs measures of dispersion in feature attribution to detect adversarial examples. 
Leveraging SimCLR, whose embedding space aligns with human perception and maintains a consistent benign-adversarial distance across various attacks, \underline{Sim}CLR encoder to \underline{cat}ch and \underline{cat}egorize adversarial attacks (SimCat) \cite{moayeri2021sample} utilizes the SimCLR distance as a robust measure for the perceptual distance, enabling generalization to unforeseen attacks. 
Drawing inspiration from the success in examining the relationship between input legitimacy and its neighborhood information for detection \cite{papernot2018deep}, latent neighborhood graph (LNG) \cite{abusnaina2021adversarial} constructs a neighborhood graph for the input and trains a graph discriminator to assess adversarial nature. 
Even if the adversarial perturbations may be challenging to discern in the pixel domain, they could induce systematic alterations in the frequency domain, rendering them detectable. Harder \etal \cite{harder2021spectraldefense} investigate detection performance by looking at both the magnitude and phase of Fourier coefficients.
\textbf{\ding{183} Difference on data distribution.} Inspired by the statistical divergence between adversarial and benign data distributions, the defense mechanism is devised to discard inputs that do not align with the benign image distribution. 
Grosse \etal \cite{grosse2017statistical} use statistical testing based on two metrics, maximum mean discrepancy (MMD) and energy distance, to distinguish adversarial distributions from benign ones. To enhance the MMD test's sensitivity to adversarial examples, Gao \etal \cite{gao2021maximum} modify the previous MMD-based detection method by using an effective kernel, optimizing kernel parameters, and addressing non-IID scenarios.
Feinman \etal \cite{feinman2017detecting} use kernel density estimation to detect points far from the benign distribution. 
MagNet \cite{meng2017magnet} hypothesizes that an autoencoder trained on a benign dataset would yield a high reconstruction error for inputs not drawn from the training data distribution, enabling MagNet to detect adversarial examples by setting a threshold for this error.
PixelDefend \cite{songpixeldefend} trains a PixelCNN as an approximation to benign probability density, using the log-likelihood of inputs as a measure to detect adversarial examples.
\textbf{\ding{184} Difference in model prediction inconsistency.} The basic idea of prediction inconsistency is to measure the sensitivity to varying decision boundaries when predicting an unknown input. This is because one adversarial example may not cross every decision boundary, while benign examples demonstrate high robustness to boundary fluctuation. 
As adversarial examples typically reside in low-confidence regions, leveraging the uncertainty output of Bayesian neural networks (BNNs) serves as a potent means of detection: an input with high predictive uncertainty implies the model has low confidence in it, thus being adversarial. Feinman \etal \cite{feinman2017detecting} approximate BNNs by using DNNs with dropout. Lightweight Bayesian refinement (LiBRe) \cite{deng2021libre} trains BNNs while balancing predictive performance, quality of uncertainty estimates, and learning efficiency.
DNNs trained with benign examples possess transformation-invariant properties, endowing them with robustness in predicting benign examples against specific input transformations. This stands in stark contrast to the vulnerability of adversarial example predictions. 
Yu \etal \cite{hu2019new} affirm an input's adversarial nature if it exhibits sensitivity to Gaussian noise, while Xu \etal \cite{xu2017feature} detect adversarial inputs through sensitivity to feature squeezing. MagNet \cite{meng2017magnet} uses autoencoder reconstruction as the image transformation. The criterion of sensitivity inconsistency detector (SID) \cite{tian2021detecting} is the sensitivity to weighted average wavelet transform (WAWT) which causes fluctuations in the highly curved region of the decision boundary.

\paragraph{Sequences of malicious queries} To avoid being fooled by the sequential attack, the inference-time defender has two choices: banning the account and giving inexact feedback, which can be shown as follows:

\textbf{1) Banning the account.} If the query from an account is detected as an intermediate state in generating adversarial examples, the defender will block this account. This kind of defense is also known as \textbf{adversarial sequential detection}.
The objective here is to differentiate between a malicious sequence of queries and a sequence for benign purposes. \underline{Pr}otecting \underline{a}gainst \underline{D}NN model stealing \underline{a}ttacks (PRADA) \cite{juuti2019prada} labels an account as malicious if its query distribution differs from that of benign accounts. Both stateful detection (SD) \cite{chen2020stateful} and Blacklight \cite{li2022blacklight} propose that the attack query sequence exhibits high self-similarity and detects the attack by identifying similar queries. To tackle persistent attackers with multiple accounts, while Blacklight \cite{li2022blacklight} stores fingerprints instead of retaining records of all prior queries like PRADA \cite{juuti2019prada} and SD \cite{chen2020stateful}, current sequential detection methods still encounter large storage overhead, warranting further improvement.

\textbf{2) Giving inexact feedback.} The defender will give inexact feedback through input, model, or output perturbation or randomization to misguide the adversary, especially for sequential query attacks. This kind of defense is also known as \textbf{perturbed prediction}.
\textbf{\ding{182} Input interference.} To disrupt the attacker's erroneous gradient estimation, the most direct approach to modifying the input is by introducing noise. Both the small noise defense (SND) method \cite{byun2022effectiveness} and the random noise defense (RND) approach \cite{qin2021random} employ Gaussian noise injection into the input. This strategy bypasses the need for adversarial training while maintaining relatively high accuracy. RND establishes theorems linking performance, noise magnitude, and the attacker's gradient step size.
\textbf{\ding{183} Model interference.} To neutralize the impact of the attacker's assault while preserving the model's predictions, Wang \etal \cite{wang2021fighting} propose interfering with the original model. They compute the gradient of model updates based on the query history, synchronizing the model and the image updates. 
\textbf{\ding{184} Ouput interference.} To preserve the predicted outcome while affecting the attacker's gradient estimation, adversarial attack on attackers (AAA) \cite{chen2022adversarial} and boundary defense (BD) \cite{aithal2022boundary} modify the soft label values, ensuring prediction accuracy. For samples with uncertain judgments, where prediction probabilities are low, BD introduces Gaussian noise to the logits. AAA leverages a designed curve to misguide the attacker, thereby extending the attacker's required time for successful attacks.

%% file: sections/section-discussion.tex
\section{Discussions}
\label{sec: discussion}

In this section, we turn our attention to the challenges and opportunities inherent in current research. Addressing these complexities is essential for advancing the adversarial machine learning field and ensuring the robustness of machine learning systems. We first discuss key challenges faced by existing defenses and highlight potential avenues for future exploration and improvement. Then we discuss some applications for these defense methods. We hope that our discussion serves as inspiration for future work in this critical area of research.

\subsection{Challenges \& opportunities}
Despite the impressive and huge growth of defense works, defenses against AML are still in their infancy. The worries and opportunities can be summarized as follows:

\paragraph{Defending against different types of attacks} 
Most existing defenses are designed for specific attacks, such as backdoor attacks \cite{wu2020adversarial}, adversarial examples \cite{lee2020adversarial}, or weight attacks \cite{ozdenizci2022improving}. In real-world applications, attackers may employ a wide range of strategies, from subtle perturbations on queried data to implanting backdoors into the model. As attackers continually develop new and sophisticated methods, defense strategies designed for specific attack types may become less effective. Attack-agnostic defenses need to be applicable in real-world, dynamic environments, where they need to account for this diversity and remain effective in the face of evolving adversarial techniques. 

On the one hand, it is imperative to investigate how defenses applied at specific stages influence one another, discerning whether the model becomes more resilient or vulnerable against attacks after implementing particular defense measures. Specifically, previous works \cite{weng2020trade,zhao2020bridging,sun2021can} have analyzed the relationship between adversarial robustness and backdoor robustness. Weng \etal \cite{weng2020trade} raised that the robustness of a DNN against adversarial examples contradicts its robustness against backdoor attacks in an empirical study. They also point out that this trade-off is not entirely detrimental to existing defenses, and a model unlearns backdoors after adversarial training in some cases \cite{wang2019neural}.
Sun \etal \cite{sun2021can} explore model robustness under different attack scenarios, and they find that forcing models to rely more on edge features can boost model robustness under adversarial examples and backdoor attacks. Zhao \etal \cite{zhao2020bridging} utilize mode connectivity to repair backdoored models and improve adversarial resistance. To conclude, robustness against different attack paradigms can be accommodated in a model, and we believe the composite defenses are a promising and important direction in the feature.
Moreover, adversarial robustness is commonly utilized to remove backdoors in backdoor defense works \cite{wang2019neural,zeng2022adversarial,mu2023progressive,zhu2023neural}. PBE \cite{mu2023progressive} argues that the adversarial examples computed by backdoor models and the poisoned samples show similarity in the feature space. Therefore, enhancing adversarial robustness can deactivate poisoned samples in the backdoored model. Nevertheless, a more comprehensive examination of the intricate relationship between these factors remains an unresolved issue, and the pursuit of a robust model that combines both adversarial robustness and backdoor robustness continues to be a valuable avenue for research.

\paragraph{Intersections between different stages of defenses}
As far as current knowledge goes, the predominant focus in existing research revolves around enhancing robustness against attacks at specific stages \cite{wu2020adversarial,wang2019neural}. However, the interconnections of defense strategies between different stages remain largely unexplored. For example, data poisoning-based backdoor attack \cite{gu2017badnets} implants a backdoor during the training process, and the backdoor is activated at the inference stage, which shows a multi-stage phenomenon for a successful attack. However, most existing backdoor defenses aim to defend against attacks at specific stages, thereby compromising the overall effectiveness of these defenses. Addressing these concerns necessitates a paradigm shift towards the adoption of multi-stage defense strategies.

\paragraph{Defense methods with practical threat models} 
There is a growing concern about the limited effectiveness of defense methods that operate under practical threat models, including considerations such as data resources \cite{nc,li2021anti,chen2022effective}, model accessibility \cite{UNICORN,Zhu_2023_ICCV}, and structural requirements \cite{zheng2022data,zheng2022pre}. Presently, defense strategies often rely on assumptions of abundant clean data, specific model architectures, or white-box accessibility. In practice, it is difficult for post-training defenders to access clean training data, and defenders at different stages may have no access to the parameters of models due to privacy concerns \cite{aiken2021neural}. These assumptions constrain their practical utility in dynamic real-world scenarios.  Future efforts could involve developing defenses capable of operating effectively with constrained access to training data or in scenarios where only out-of-distribution datasets are available. The importance of black-box defenses becomes particularly pronounced in situations where the internal workings of the model remain partially unknown. Exploring defense methods based on practical threat models is pivotal for enhancing the applicability and robustness of defense mechanisms in the face of evolving and sophisticated adversarial threats, paving the way for more resilient machine learning systems in practical, less controlled environments.

\paragraph{Defense against different learning paradigm scenarios} An important concern is defenses across different learning paradigms, particularly in the context of the prevailing importance of semi-supervised \cite{li2022semi} and self-supervised learning \cite{naseer2020self,moayeri2021sample,feng2023detecting}. As these alternative paradigms, exemplified by their prominent role in the development of foundation models \cite{Bansal_2023_ICCV,liang2023badclip}, gain prominence, the current dearth of defenses tailored to safeguard them against adversarial threats becomes more pronounced. By extending the focus beyond traditional supervised learning, researchers can explore more dimensions of defense strategies that align with emerging paradigms, ensuring a more comprehensive and versatile defense against adversarial threats in evolving and different learning environments.

\paragraph{Defending against multi-modality scenario} 
A critical concern lies in the scarcity of approaches tailored for multi-modality scenarios \cite{baltruvsaitis2018multimodal}. The worry arises from the diverse characteristics and data distributions inherent in different modalities such as computer vision (CV), natural language processing (NLP), audio, and multi-modality models. Current defense strategies, often designed for a specific modality, may struggle to generalize effectively and fail in multi-modality scenarios \cite{liang2023badclip,walmer2022dual}, leaving certain applications vulnerable to adversarial exploits. Even if there are already some defenses \cite{Bansal_2023_ICCV} against multi-modality scenarios, these methods can easily fail in the face of endless multi-modal attacks. Addressing this gap requires future efforts to draw insights from defenses in various domains and prioritize the development of defenses capable of adapting across multi-modality application areas, ensuring resilience against adversarial threats regardless of the specific domain.

\paragraph{Developing more comprehensive evaluation metrics} Another significant concern is the deficiency of comprehensive evaluation metrics. The prevailing focus on attack success rate, model accuracy, and defense effectiveness rating \cite{Zhu_2023_ICCV} overlooks essential aspects, particularly the robustness and the temporal and computational costs \cite{jia2022certified,weber2023rab} associated with defense mechanisms, which are also crucial for real-world applications. Future research should prioritize the development of comprehensive metrics that encompass both computational dimensions and defending effects against different aspects of attacks, offering a more encompassing evaluation of defense strategies. Additionally, establishing a comprehensive evaluation system is instrumental in facilitating the development of effective defense mechanisms.

\subsection{Applications}
Although defense methods are primarily designed for protecting neural networks from different types of attacks, these methods can be used beyond the scope of model robustness and have some applications in diverse contexts.  This section provides a concise overview of recent works exploring the broader applications of defense methods. Notably, our focus here is on the application of backdoor defense and adversarial defense, as weight defenses currently lack specific applications and are thus excluded from this discussion.

\paragraph{Backdoor defense for evaluating copyright protection} Recent works \cite{zhong2023copyright,xu2023watermarking} have explored the use of backdoor attacks for ownership verification and copyright protection for pre-trained models and datasets. The idea is to embed a backdoor watermark into the model or the data and use it to prove the ownership or authenticity of the model or the data.
Backdoor watermarking has been shown effective for watermarking graph neural networks \cite{xu2023watermarking}, generative adversarial networks \cite{zhong2023copyright}, large language models \cite{peng2023you}, and dataset copyright protection \cite{guo2023domain}.
Accordingly, backdoor detection and defense strategies can be utilized to evaluate the effectiveness of these copyright protection methods. These methods aim to determine whether the model has a watermark or evaluate the robustness of watermarks by removing them. For example, Liu \etal \cite{liu2021removing} propose WILD to remove the watermarks of models with limited training data and find that the output model can perform the same as models trained from scratch without watermarks injected. Aiken \etal \cite{aiken2021neural} propose a method to effectively remove black-box backdoor watermarks with no prior knowledge of the structure of networks. They show that the robustness of the watermark is significantly weaker than the original claims.
Some standard backdoor defense methods \cite{liu2018fine,wu2021adversarial} are also applied to evaluate the robustness of backdoor-based watermarks \cite{guo2023domain}.

\paragraph{Defense against adversarial examples} 
Firstly, adversarial training can be applied for privacy preservation \cite{pittaluga2019learning}. Deep neural networks necessitate substantial amounts of data for optimal training, and the training data may contain personally identifiable information and other sensitive details. Consequently, privacy preservation is paramount, aiming to safeguard the sensitive information inherent in data through the implementation of specialized training strategies. One such strategy is adversarial training, which aims to inhibit inference of chosen private attributes and enhance the model's resilience against potential privacy leakage via adversarial training \cite{pittaluga2019learning,ding2020privacy,hsieh2021netfense}.

Second, adversarial training can be used to improve out-of-distribution (OOD) generalization \cite{yi2021improved}. OOD data are data that do not follow the same distribution as the training data and may cause the neural network to produce inconsistent and incomprehensible predictions. Improving the OOD generalization of neural networks is a challenging and important problem, especially in safety-critical and security-sensitive domains. Some works have pointed out the connection between input-robustness and OOD generalization, and show that adversarial training can improve the OOD generalization of neural networks \cite{yi2021improved, xin2023connection}. For example, Xin \etal \cite{yi2021improved} theoretically find that a robust model also generalizes well on OOD data, and they improve OOD generalization by utilizing adversarial training. Adversarial training can also be utilized for privacy-preserving in visual recognition. DAT \cite{xin2023connection} studies the similarity between invariant risk minimization and proposes domain-wise adversarial training (DAT) to alleviate distribution shift by domain-specific perturbations.

%% file: sections/section-conclusion.tex
\section{Conclusion}
\label{sec:conclusion}

Adversarial vulnerabilities in machine learning models pose significant threats to the integrity, availability, and confidentiality of these systems. This survey addresses the overarching field of adversarial machine learning, focusing on three primary attack methodologies: backdoor attacks, weight attacks, and adversarial examples, each targeting distinct stages of the machine learning life-cycle. Our contribution lies in proposing a unified perspective on AML defenses based on the machine learning system's life-cycle, which aligns diverse defenses with various attacks at different stages, facilitating comparisons and analyses within each stage. We also present taxonomies to categorize different defenses in different stages, and provide an extensive survey of existing defenses. Besides, we identify open challenges and future directions for the development of more effective and robust defenses throughout the machine learning system's life-cycle. We believe that our unified perspective and comprehensive survey will guide future research in developing more resilient defenses against adversarial attacks in machine learning.
